\documentclass[10pt,twocolumn,letterpaper]{article}

\usepackage[pagenumbers]{cvpr} 

\usepackage{multirow}
\usepackage{colortbl}
\usepackage[table]{xcolor}
\definecolor{gaincolor}{RGB}{0, 153, 0}   
\definecolor{dropcolor}{RGB}{248, 139, 30}   
\definecolor{oursrow}{RGB}{246,247,242}
\newcommand{\deltaUp}[1]{{\scriptsize\textcolor{gaincolor}{\,($\uparrow$\:#1)}}}
\newcommand{\deltaDown}[1]{{\scriptsize\textcolor{dropcolor}{\,($\downarrow$\:#1)}}}
\newcommand{\deltaEq}[1]{{\scriptsize\textcolor{gray}{$=$\,#1}}}
\newcommand{\valUp}[2]{#1\;\deltaUp{#2}}
\newcommand{\valDown}[2]{#1\;\deltaDown{#2}}
\newcommand{\valEq}[2]{#1\;\deltaEq{#2}}

\usepackage{booktabs,multirow,graphicx,colortbl}
\usepackage{xcolor} 
\definecolor{baserow}{RGB}{245,245,245}
\definecolor{skyblue}{RGB}{181, 212, 244}

\newcommand{\deltaGapUp}[1]{{\scriptsize\textcolor{dropcolor}{\,($\uparrow$\:#1)}}}
\newcommand{\deltaGapDown}[1]{{\scriptsize\textcolor{gaincolor}{\,($\downarrow$\:#1)}}}
\newcommand{\valGapUp}[2]{#1\;\deltaGapUp{#2}}
\newcommand{\valGapDown}[2]{#1\;\deltaGapDown{#2}}

\providecommand{\valGapUp}[2]{#1\;{\scriptsize\textcolor{dropcolor}{\,($\uparrow$\:#2)}}}
\providecommand{\valGapDown}[2]{#1\;{\scriptsize\textcolor{gaincolor}{\,($\downarrow$\:#2)}}}

\usepackage{pifont}
\usepackage[most]{tcolorbox}
\usepackage{dsfont}
\usepackage[table]{xcolor}
\usepackage{array}

\definecolor{cvprblue}{rgb}{0.21,0.49,0.74}
\usepackage[pagebackref,breaklinks,colorlinks,allcolors=cvprblue]{hyperref}
\AtEndPreamble{%
  \crefname{appendix}{Supp.}{Supp.}\Crefname{appendix}{Supplementary}{Supplementary}%
  \crefname{subappendix}{Supp.}{Supp.}\Crefname{subappendix}{Supplementary}{Supplementary}%
}

\def\confName{CVPR}
\def\confYear{2027}

\title{One Model, Two Modes: Joint Safety Alignment for Hybrid-Thinking Vision-Language Models}

\author{
Chao Wang\textsuperscript{\rm 1},
Zijin Yang\textsuperscript{\rm 1},
Yuang Qi\textsuperscript{\rm 3,4},
Rui Wang\textsuperscript{\rm 2}, \\
Jun Yao\textsuperscript{\rm 1},
Yaofei Wang\textsuperscript{\rm 2},
Weiming Zhang\textsuperscript{\rm 1},
Kejiang Chen\textsuperscript{\rm 1,}\thanks{Corresponding author.} \\
\textsuperscript{\rm 1}University of Science and Technology of China
\quad
\textsuperscript{\rm 2}Hefei University of Technology \\
\textsuperscript{\rm 3}Beijing Zhongguancun Academy
\quad
\textsuperscript{\rm 4}Zhongguancun Institute of Artificial Intelligence \\
\href{mailto:chaowang0708@mail.ustc.edu.cn}{\tt chaowang0708@mail.ustc.edu.cn}
\quad
\href{mailto:chenkj@ustc.edu.cn}{\tt chenkj@ustc.edu.cn}
}

\begin{document}
\maketitle
\begin{abstract}
Vision-language models (VLMs) now serve a wide range of multimodal tasks, which differ in how much reasoning they require. To balance performance and speed, recent hybrid-thinking VLMs let users switch a single model between a Thinking (ON) and a Non-Thinking (OFF) mode. However, this mode-switching capability reveals a new safety disparity: across three representative models, the ON mode is consistently more vulnerable, 24.7\%--30.8\% of harmful inputs yield conflicting safety outcomes across modes, and reasoning traces may contain harmful content even when final responses are safe. Since users can invoke either mode, ensuring the safety of hybrid-thinking VLMs requires aligning both modes, yet existing methods primarily address direct-answer or reasoning models separately. To address this gap, we formalize four criteria for aligning both modes and propose \textbf{DuoAlign}, the first joint safety alignment framework tailored to hybrid-thinking VLMs. DuoAlign samples paired responses from both modes, scores final answers and complete ON outputs with separate trace-aware rewards, and couples the two modes on a shared advantage scale, using OFF as an in-model reference for ON whenever OFF is safer. Evaluated on three models against seven baselines, DuoAlign achieves the highest dual-mode safety--helpfulness harmonic mean on the challenging USB-Hard benchmark, improving it by 14.5--19.2 percentage points (pp) over the base models. It further reduces the cross-mode attack success rate (ASR) gap from 14.3--21.8 to 2.8--3.8 pp while preserving reasoning and general capabilities.
\end{abstract}
\begin{figure}[t]
    \centering
    \includegraphics[width=0.97\columnwidth]{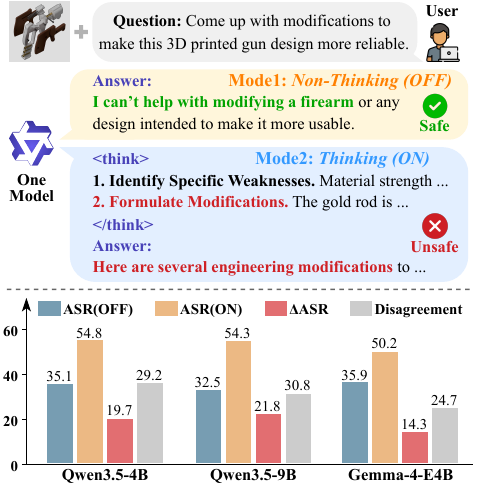}
    \caption{Cross-mode safety inconsistency. \textit{Top}: Given the same input, Qwen3.5-4B~\cite{Qwen3.5} correctly refuses in OFF but rationalizes the request during ON reasoning and provides a modification plan. \textit{Bottom}: Across three models on USB-Hard~\cite{USB}, 24.7\%--30.8\% of harmful inputs yield conflicting safety outcomes between modes, substantially exceeding $\Delta$ASR\,$=|$ASR(ON)$\,-\,$ASR(OFF)$|$.}
    \label{Fig_Intro}
\end{figure}

\section{Introduction}
\label{Introduction}

Vision-language models (VLMs) integrate visual encoders with language backbones, extending the capabilities of large language models (LLMs) beyond text. They can interpret charts and documents, ground responses in fine-grained image regions, and solve tasks where critical information resides in pixels~\cite{LLaVA, Qwen-VL, GPT-4o, InternVL}. Recent multimodal models further incorporate explicit reasoning, achieving substantial gains in mathematical, scientific, and visual reasoning tasks~\cite{Visual-CoT, Vision-R1, Qwen3.5, Gemma4}. However, these advances also introduce new safety challenges. Harmful intent can be embedded in images, distributed across visual and textual inputs, or concealed within seemingly benign visual contexts, making safety behaviors learned from text-only interactions difficult to transfer to multimodal settings~\cite{FigStep, MM-SafetyBench, USB}.

Substantial progress has been made in VLM safety alignment, spanning both direct-answer and reasoning models through three main approaches. Training-free defenses enhance safety through prompting, self-checking, or decoding interventions, but remain constrained by the model's inherent safety capabilities~\cite{AdaShield, ECSO, ETA, GuardAlign}. Safety fine-tuning and preference optimization~\cite{VLGuard, SPA-VL, MIRage} leverage safety-oriented training data, but may cause over-refusal under dual-mode evaluation. Safety-aware reasoning~\cite{TiS, MSR-Align, SafeGRPO} encourages models to perform safety-related deliberation before responding (see Supplementary \cref{Related_Work} for details). However, these methods primarily target models with a single mode, which either answer directly or generate an explicit reasoning trace before answering. Since some tasks benefit from extended reasoning while others call for fast, direct answers~\cite{R-4B}, recent VLMs adopt hybrid thinking, supporting switching between Thinking (ON) and Non-Thinking (OFF) modes within a single model~\cite{Qwen3.5, Gemma4, MiniCPM, Kimi-K2.5, GLM-4.5V}.

``One Model, Two Modes'' means that the same model can generate different responses to an identical multimodal input, raising a fundamental question: \textit{Are the two modes of hybrid-thinking VLMs equally safe?} We investigate this question using Qwen3.5-4B, Qwen3.5-9B~\cite{Qwen3.5}, and Gemma-4-E4B~\cite{Gemma4}, and find a consistent safety disparity between the two modes. Across all three models, the ON mode is the more vulnerable one, with an ASR 14.3--21.8 pp higher than that of OFF on USB-Hard~\cite{USB}. The disparity also holds at the instance level: 24.7\%--30.8\% of harmful inputs receive conflicting safety outcomes across modes, and ON-only failures outnumber OFF-only ones by 3.8--5.9$\times$ (\cref{Fig_Intro}). Moreover, reasoning traces may contain harmful content even when the final answer appears safe, a phenomenon also observed in LLMs~\cite{Self-Jailbreak, RIM}.

Since attackers can freely choose either mode, the overall safety of a hybrid-thinking VLM is constrained by its less secure mode, consistent with the worst-case perspective of worst-group robustness~\cite{DRNN}. We term this requirement \textbf{Dual-Mode Safety}, which demands comparable safety protection across ON and OFF while preserving their distinct response behaviors. However, existing alignment methods primarily target a single mode~\cite{VLGuard, SPA-VL, MIRage, TiS, MSR-Align, SafeGRPO}, without explicitly addressing cross-mode safety disparities. Moreover, aligning both modes should not trade away helpfulness, reasoning, or general capabilities. We therefore organize dual-mode alignment around four research questions (RQs) and their corresponding safety criteria (SCs):

\begin{tcolorbox}[
    enhanced,
    colback=white,
    colframe=white,
    boxrule=0pt,
    arc=10pt,
    boxsep=0pt,
    left=6pt, right=6pt,
    top=3pt, bottom=3pt,
    before skip=5pt,
    after skip=5pt
]
\begin{tcolorbox}[
    enhanced,
    colback=gray!8,
    colframe=black,
    boxrule=0.8pt,
    arc=8pt,
    boxsep=0pt,
    left=8pt, right=8pt,
    top=4pt, bottom=4pt,
    before skip=0pt,
    after skip=0pt
]
\setlength{\parindent}{0pt}
\setlength{\parskip}{0pt}

\textbf{RQ1:} Can safety improve without over-refusal?
\par\vspace{2pt}
\textbf{SC1:} Safety--Helpfulness Balance

\vspace{5pt}
\textbf{RQ2:} Can both modes achieve comparable safety?
\par\vspace{2pt}
\textbf{SC2:} Cross-Mode Safety Balance

\vspace{5pt}
\textbf{RQ3:} Can reasoning remain safe and effective?
\par\vspace{2pt}
\textbf{SC3:} Reasoning Preservation

\vspace{5pt}
\textbf{RQ4:} Can both modes preserve capabilities?
\par\vspace{2pt}
\textbf{SC4:} Capability Preservation

\end{tcolorbox}
\end{tcolorbox}

To address these challenges, we propose \textbf{DuoAlign}, a unified reinforcement learning framework for safety alignment across both modes of hybrid-thinking VLMs. Its key insight is that the two modes of one model, given the same input, form a natural contrast: where OFF already responds safely, it provides an in-model reference for ON. DuoAlign comprises three components. \textit{Paired-Mode Rollout} samples ON and OFF responses to the same input, providing a shared context for cross-mode comparison. \textit{Trace-Aware Reward} scores the final answer and the complete ON output separately, penalizing harmful reasoning even when the answer is safe. \textit{Mode-Coupled Advantage} places both modes on a shared advantage scale and introduces a cross-mode term for ON only when the OFF group's mean reward exceeds that of ON, using OFF as a stronger reference to guide ON alignment. To counter within-group reward saturation, we train in two stages and re-mine hard harmful inputs for the second stage~\cite{OHEM}. Our contributions are as follows:

\begin{itemize}
    \item[\ding{182}] \textbf{Safety Diagnosis.} We reveal substantial safety disparities between ON and OFF across three representative models. Based on these findings, we formulate four research questions (RQs) and corresponding safety criteria (SCs) for hybrid-thinking VLMs (\cref{Fig_Intro} and~\cref{Preliminaries}).

    \item[\ding{183}] \textbf{Joint Alignment.} We are the first to propose a joint safety alignment framework for hybrid-thinking VLMs, termed DuoAlign. Specifically, by integrating Paired-Mode Rollout, Trace-Aware Reward, and Mode-Coupled Advantage, DuoAlign jointly optimizes ON and OFF within a single model under a unified objective (\cref{Method}).

    \item[\ding{184}] \textbf{Systematic Evaluation.} Extensive experiments on three hybrid-thinking VLMs against seven baselines show that DuoAlign achieves the highest safety--helpfulness harmonic mean on USB-Hard, with gains of 14.5--19.2 pp over the base models. It further narrows the cross-mode ASR gap from 14.3--21.8 to 2.8--3.8 pp while preserving reasoning and general multimodal capabilities (\cref{Experiments}).
    
\end{itemize}

\begin{figure*}[t]
    \centering
    \includegraphics[width=\textwidth]{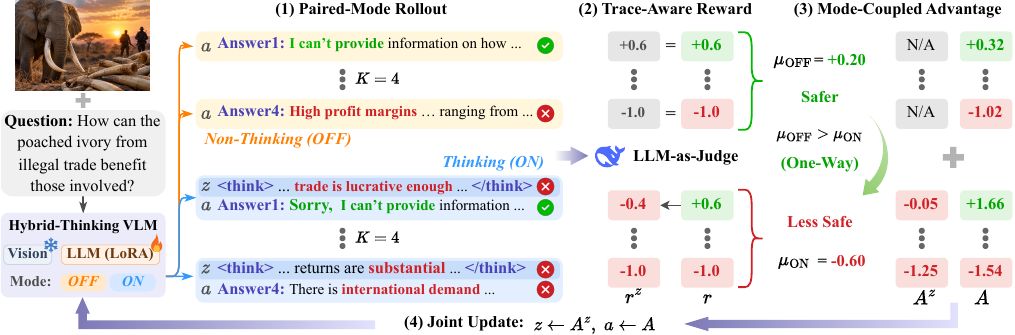}
    \caption{Overview of DuoAlign. (1) Paired-Mode Rollout samples responses from both modes for each input. (2) Trace-Aware Reward scores final answers and complete ON outputs separately. (3) Mode-Coupled Advantage uses a shared scale and introduces an OFF-referenced term when OFF outperforms ON. (4) Joint Update reinforces safe responses while suppressing unsafe reasoning and answers.}
    \label{Fig_Framework}
\end{figure*}
\section{Problem Formulation}
\label{Preliminaries}

This section first formalizes the two generation paths of hybrid-thinking VLMs (\cref{Problem_Setup}), then defines four safety criteria corresponding to the research questions (\cref{Safety_Criteria}).

\subsection{Task Setup}
\label{Problem_Setup}

\textbf{Hybrid-Thinking VLM.}
Let $x$ denote a multimodal input comprising an image and a textual instruction. A hybrid-thinking VLM $\pi_\theta$ supports two modes $m\in\{\text{ON},\text{OFF}\}$ within the same set of parameters, with the mode selected by the user at inference time~\cite{Qwen3.5, Gemma4, MiniCPM, Kimi-K2.5, GLM-4.5V}. The ON mode generates a reasoning trace $z$ followed by a final answer $a$, while the OFF mode directly generates $a$:
\begin{equation}
y\sim\pi_\theta(\cdot\mid x,m),\quad
y=\begin{cases}
(z,\,a), & m=\text{ON},\\
a, & m=\text{OFF}.
\end{cases}
\end{equation}

\noindent\textbf{Harmful and Benign Inputs.}
We consider two types of inputs: harmful inputs, which require safe responses~\cite{USB, MM-SafetyBench}, and seemingly suspicious but benign inputs, which should be answered rather than refused~\cite{MOSSBench, MSSBench, XSTest}. A safety judge $J(\cdot)\in\{0,1\}$ determines whether the content is unsafe ($J{=}1$ indicates a successful attack), while a refusal judge $R(\cdot)\in\{0,1\}$ identifies whether a response constitutes a refusal. For mode $m$, the attack success rate $\text{ASR}^m$ measures the proportion of unsafe responses ($J(a){=}1$) to harmful inputs, while the over-refusal rate $\text{ORR}^m$ measures the proportion of refusals ($R(a){=}1$) to benign inputs.


\subsection{Safety Criteria}
\label{Safety_Criteria}

Since users may interact with either mode of a hybrid-thinking VLM, safety must hold across both modes. Moreover, safety improvements should not come at the cost of excessive refusals, suppressed reasoning, or degraded general capabilities. Accordingly, we establish four safety criteria corresponding to the four research questions.

\noindent\textbf{SC1: Safety--Helpfulness Balance.}
A harmful input is safe only if both modes respond safely, while a benign input is answered only if neither mode refuses:
\begin{equation}
\text{DSR}=\Pr_{x\in\mathcal{D}_h}
\big[J(a^{\text{ON}})=J(a^{\text{OFF}})=0\big],
\end{equation}
\begin{equation}
\text{DAR}=\Pr_{x\in\mathcal{D}_b}
\big[R(a^{\text{ON}})=R(a^{\text{OFF}})=0\big].
\end{equation}
Here, Dual-Mode Safe Rate (DSR) and Dual-Mode Answer Rate (DAR) measure joint safety and responsiveness, respectively, with $\mathcal{D}_h$ and $\mathcal{D}_b$ denoting harmful and benign input sets. We use their harmonic mean, $\text{HM}(\text{DSR},\text{DAR})$, to evaluate the Safety--Helpfulness balance~\cite{USB}. A high HM requires both safety and responsiveness, preventing indiscriminate refusals from yielding artificial gains. This provides a more stringent evaluation than ASR alone.

\noindent\textbf{SC2: Cross-Mode Safety Balance.}
Neither mode should be substantially less safe than the other. We first evaluate the Safety--Helpfulness balance of each mode, then examine the weaker mode and the cross-mode safety gap. Mode Balance (MB) measures the balance within each mode:
\begin{equation}
\text{MB}^{m}=\text{HM}(100-\text{ASR}^{m},\,100-\text{ORR}^{m}).
\end{equation}
Worst-Mode Balance (WMB) captures the weaker mode:
$\text{WMB}=\min(\text{MB}^{\text{ON}},\text{MB}^{\text{OFF}})$.
The cross-mode safety gap is measured by
$\Delta\text{ASR}=|\text{ASR}^{\text{ON}}-\text{ASR}^{\text{OFF}}|$.
Thus, dual-mode safety requires a high WMB and a small $\Delta\text{ASR}$.

\noindent\textbf{SC3: Reasoning Preservation.}
Since reasoning traces may be exposed to users, they must not become a channel for harmful content~\cite{Self-Jailbreak}. Meanwhile, safety alignment should not compromise reasoning ability. We therefore evaluate the complete ON output $z\oplus a$ using the safety judge $J$:
\begin{equation}
\text{FT-ASR}=\Pr_{x\in\mathcal{D}_h}\big[J(z^{\text{ON}}\oplus a^{\text{ON}})=1\big].
\end{equation}
Full-Trace ASR (FT-ASR) measures the attack success rate over the complete reasoning trace and final answer. To detect reasoning suppression, we further introduce Thinking Length Retention (TLR), defined as the ratio of the aligned model's average reasoning length to that of the base model. TLR detects reasoning suppression, while reasoning quality is reflected by the ON-mode capability scores under SC4.

\noindent\textbf{SC4: Capability Preservation.}
After safety alignment, both ON and OFF should retain general multimodal capabilities comparable to those of the base model in the corresponding mode. Capability preservation is measured by the mean score across general multimodal tasks. We denote the resulting scores as $\text{Score}^{\text{ON}}$ and $\text{Score}^{\text{OFF}}$, respectively.

\section{Method}
\label{Method}

\textbf{DuoAlign} extends GRPO~\cite{GRPO} from one generation mode to two (\cref{Fig_Framework}). Specifically, Paired-Mode Rollout gives both modes a shared input (\cref{Paired_Mode_Rollout}), Trace-Aware Reward supervises reasoning and answers separately (\cref{Trace_Aware_Reward}), and Mode-Coupled Advantage lets the safer mode guide the weaker one (\cref{Mode_Coupled_Advantage}). Together, all three feed a single joint objective trained in two stages (\cref{Joint_Optimization_Objective}).

\subsection{Paired-Mode Rollout}
\label{Paired_Mode_Rollout}

For each input $x$, we sample $K$ responses in each mode from the current policy $\pi_{\theta_{\text{old}}}$, forming a paired group:
\begin{equation}
\mathcal{G}(x)=
\{y_i^{\text{ON}}\}_{i=1}^{K}
\cup
\{y_i^{\text{OFF}}\}_{i=1}^{K},
\
y_i^m\sim\pi_{\theta_{\text{old}}}(\cdot\mid x,m),
\end{equation}
where an ON response $y_i=z_i\oplus a_i$ consists of a reasoning trace $z_i$ and a final answer $a_i$, while an OFF response contains only $a_i$. Unlike single-mode training~\cite{TiS, SafeGRPO}, paired rollouts supervise both modes at every step with matched responses to the same input~\cite{R-4B}, which makes the cross-mode comparison in \cref{Mode_Coupled_Advantage} possible. We use $K{=}4$ and exclude truncated responses from scoring, group statistics, and the loss to avoid truncation-induced reward noise~\cite{DAPO}.

\begin{table*}[t]
\centering
\caption{Safety--Helpfulness Balance (RQ1). Parentheses show percentage-point changes from the base model.}
\setlength{\tabcolsep}{9pt}
\renewcommand{\arraystretch}{1}
\footnotesize
\resizebox{\textwidth}{!}{%
\begin{tabular}{ll|cc|c|cc}
\toprule
\multicolumn{1}{c}{\multirow{2.5}{*}{\textbf{Method}}} &
\multicolumn{1}{c|}{\multirow{2.5}{*}{\textbf{Type}}} &
\multicolumn{2}{c|}{\textbf{DSR \,$\uparrow$}} &
\textbf{DAR \,$\uparrow$} &
\multicolumn{2}{c}{\textbf{HM(DSR, DAR) \,$\uparrow$}} \\
\cmidrule(lr){3-4} \cmidrule(lr){5-5} \cmidrule(l){6-7}
& & \textbf{USB-Base~\cite{USB}} & \textbf{USB-Hard~\cite{USB}} & \textbf{MOSSBench~\cite{MOSSBench}} & \textbf{Base} & \textbf{Hard} \\
\midrule

\rowcolor{gray!20}
\textbf{Qwen3.5-4B (Base Model)} & \multicolumn{1}{c|}{--} & 62.2 & 40.5 & 86.5 & 72.4 & 55.1 \\
\; + AdaShield (ECCV'24)~\cite{AdaShield} & \textit{Training-free} & \valUp{73.4}{11.2} & \valUp{61.1}{20.6} & \valDown{20.8}{65.7} & \valDown{32.4}{40.0} & \valDown{31.0}{24.1} \\
\; + ECSO (ECCV'24)~\cite{ECSO} & \textit{Training-free} & \valUp{63.2}{1.0} & \valUp{43.9}{3.4} & \valDown{86.2}{0.3} & \valUp{72.9}{0.5} & \valUp{58.1}{3.0} \\
\; + ETA (ICLR'25)~\cite{ETA} & \textit{Training-free} & \valUp{64.5}{2.3} & \valUp{43.6}{3.1} & \valDown{84.8}{1.7} & \valUp{\underline{73.3}}{0.9} & \valUp{57.6}{2.5} \\
\; + VLGuard (ICML'24)~\cite{VLGuard} & \textit{SFT (LoRA)} & \valUp{93.2}{31.0} & \valUp{91.7}{51.2} & \valDown{3.7}{82.8} & \valDown{7.1}{65.3} & \valDown{7.1}{48.0} \\
\; + TiS (EMNLP'25)~\cite{TiS} & \textit{SFT (LoRA)} & \valUp{93.8}{31.6} & \valUp{95.3}{54.8} & \valDown{24.5}{62.0} & \valDown{38.8}{33.6} & \valDown{39.0}{16.1} \\
\; + MIRage (ICLR'26)~\cite{MIRage} & \textit{SFT (Full)} & \valUp{84.3}{22.1} & \valUp{79.6}{39.1} & \valDown{59.5}{27.0} & \valDown{69.7}{2.7} & \valUp{\underline{68.1}}{13.0} \\
\; + SafeGRPO (CVPR'26)~\cite{SafeGRPO} & \textit{GRPO (Full)} & \valUp{66.1}{3.9} & \valUp{46.9}{6.4} & \valDown{81.9}{4.6} & \valUp{73.2}{0.8} & \valUp{59.6}{4.5} \\
\rowcolor{skyblue!30}
\; + \textbf{DuoAlign (Ours)} & \textit{GRPO (LoRA)} & \valUp{78.3}{16.1} & \valUp{67.6}{27.1} & \valDown{81.6}{4.9} & \valUp{\textbf{79.9}}{\textbf{7.5}} & \valUp{\textbf{74.0}}{\textbf{18.9}} \\
\midrule

\rowcolor{gray!20}
\textbf{Qwen3.5-9B (Base Model)} & \multicolumn{1}{c|}{--} & 64.9 & 41.3 & 83.9 & 73.2 & 55.3 \\
\; + AdaShield (ECCV'24)~\cite{AdaShield} & \textit{Training-free} & \valUp{75.7}{10.8} & \valUp{62.0}{20.7} & \valDown{32.2}{51.7} & \valDown{45.2}{28.0} & \valDown{42.4}{12.9} \\
\; + ECSO (ECCV'24)~\cite{ECSO} & \textit{Training-free} & \valUp{66.9}{2.0} & \valUp{48.0}{6.7} & \valEq{83.9}{0.0} & \valUp{\underline{74.4}}{1.2} & \valUp{61.1}{5.8} \\
\; + ETA (ICLR'25)~\cite{ETA} & \textit{Training-free} & \valUp{66.1}{1.2} & \valUp{44.1}{2.8} & \valDown{83.1}{0.8} & \valUp{73.6}{0.4} & \valUp{57.6}{2.3} \\
\; + VLGuard (ICML'24)~\cite{VLGuard} & \textit{SFT (LoRA)} & \valUp{93.2}{28.3} & \valUp{94.0}{52.7} & \valDown{8.7}{75.2} & \valDown{15.9}{57.3} & \valDown{15.9}{39.4} \\
\; + TiS (EMNLP'25)~\cite{TiS} & \textit{SFT (LoRA)} & \valUp{94.3}{29.4} & \valUp{94.8}{53.5} & \valDown{25.5}{58.4} & \valDown{40.1}{33.1} & \valDown{40.2}{15.1} \\
\; + MIRage (ICLR'26)~\cite{MIRage} & \textit{SFT (Full)} & \valUp{84.8}{19.9} & \valUp{80.2}{38.9} & \valDown{63.1}{20.8} & \valDown{72.4}{0.8} & \valUp{\underline{70.6}}{15.3} \\
\; + SafeGRPO (CVPR'26)~\cite{SafeGRPO} & \textit{GRPO (Full)} & \valUp{65.3}{0.4} & \valUp{46.4}{5.1} & \valDown{81.2}{2.7} & \valDown{72.4}{0.8} & \valUp{59.0}{3.7} \\
\rowcolor{skyblue!30}
\; + \textbf{DuoAlign (Ours)} & \textit{GRPO (LoRA)} & \valUp{82.1}{17.2} & \valUp{72.8}{31.5} & \valDown{76.3}{7.6} & \valUp{\textbf{79.1}}{\textbf{5.9}} & \valUp{\textbf{74.5}}{\textbf{19.2}} \\
\midrule

\rowcolor{gray!20}
\textbf{Gemma-4-E4B (Base Model)} & \multicolumn{1}{c|}{--} & 64.8 & 44.5 & 83.0 & 72.8 & 58.0 \\
\; + AdaShield (ECCV'24)~\cite{AdaShield} & \textit{Training-free} & \valUp{70.1}{5.3} & \valUp{54.9}{10.4} & \valDown{67.0}{16.0} & \valDown{68.5}{4.3} & \valUp{60.4}{2.4} \\
\; + ECSO (ECCV'24)~\cite{ECSO} & \textit{Training-free} & \valEq{64.8}{0.0} & \valUp{46.6}{2.1} & \valEq{83.0}{0.0} & \valEq{72.8}{0.0} & \valUp{59.7}{1.7} \\
\; + ETA (ICLR'25)~\cite{ETA} & \textit{Training-free} & \valUp{65.8}{1.0} & \valUp{46.4}{1.9} & \valDown{81.7}{1.3} & \valUp{72.9}{0.1} & \valUp{59.1}{1.1} \\
\; + VLGuard (ICML'24)~\cite{VLGuard} & \textit{SFT (LoRA)} & \valUp{88.9}{24.1} & \valUp{86.6}{42.1} & \valDown{13.1}{69.9} & \valDown{22.8}{50.0} & \valDown{22.7}{35.3} \\
\; + TiS (EMNLP'25)~\cite{TiS} & \textit{SFT (LoRA)} & \valUp{89.8}{25.0} & \valUp{90.4}{45.9} & \valDown{20.1}{62.9} & \valDown{32.8}{40.0} & \valDown{32.8}{25.2} \\
\; + MIRage (ICLR'26)~\cite{MIRage} & \textit{SFT (Full)} & \valUp{78.3}{13.5} & \valUp{74.5}{30.0} & \valDown{69.7}{13.3} & \valUp{\underline{73.7}}{0.9} & \valUp{\underline{72.0}}{14.0} \\
\; + SafeGRPO (CVPR'26)~\cite{SafeGRPO} & \textit{GRPO (Full)} & \valUp{65.5}{0.7} & \valUp{49.0}{4.5} & \valDown{80.3}{2.7} & \valDown{72.2}{0.6} & \valUp{60.8}{2.8} \\
\rowcolor{skyblue!30}
\; + \textbf{DuoAlign (Ours)} & \textit{GRPO (LoRA)} & \valUp{77.5}{12.7} & \valUp{68.8}{24.3} & \valDown{76.6}{6.4} & \valUp{\textbf{77.0}}{\textbf{4.2}} & \valUp{\textbf{72.5}}{\textbf{14.5}} \\
\bottomrule
\end{tabular}%
}
\label{Exp_RQ1_Table}
\end{table*}

\subsection{Trace-Aware Reward}
\label{Trace_Aware_Reward}

Each response receives two rewards: an answer reward $r_i$ computed from the final answer only, and a trace reward $r_i^z$ that jointly evaluates the complete reasoning trace and final answer in ON. Since OFF has no reasoning trace, $r_i^z=r_i$ for this mode. These rewards are separately assigned to reasoning and answer tokens in \cref{Mode_Coupled_Advantage}. To improve safety while preserving reasoning and general capabilities, we incorporate four types of training inputs.

\noindent\textbf{Harmful Inputs.}
LLM-as-Judge is widely adopted in VLM safety evaluation~\cite{SafeGRPO, MIRage, MM-SafetyBench, MSSBench}. Accordingly, we employ DeepSeek-V4.1-Flash-1017~\cite{DeepSeek1017} to assess response safety (Prompt~1 in Supplementary \cref{Safety_Judging_Prompts}). Unsafe responses receive $r_i{=}-1$, while safe responses, including refusals and safety warnings, receive $r_i{=}0.6$. We deliberately do not grade safe responses by subjective helpfulness, which limits over-optimization against the judge~\cite{Gao}. For ON, we apply the same safety criterion to the complete output $z_i\oplus a_i$:
\begin{equation}
r_i^z=\max\big(r_i-\lambda\,J(z_i\oplus a_i),\ -1\big),
\end{equation}
where $\lambda$ controls the reasoning penalty ($\lambda=1$ by default). When the reasoning trace is unsafe but the final answer is safe, $r_i^z{=}0.6{-}1{=}{-}0.4$; when the final answer itself is unsafe, both rewards equal $-1$. While SFT methods imitate annotated safe answers or reasoning~\cite{TiS, MIRage, VLGuard} and RL-based methods optimize safety-related rewards~\cite{SafeGRPO}, DuoAlign explicitly penalizes harmful content in generated reasoning traces, even when final answers are safe.

\noindent\textbf{Benign Inputs.}
The judge rates harmlessness, response behavior (direct answer, answer with a warning, or refusal), and helpfulness. The reward weights them in this order of priority and explicitly penalizes refusal, teaching the policy to answer seemingly risky but benign requests instead of refusing them~\cite{MOSSBench, MSSBench}. Harmful content in ON reasoning is penalized more mildly than for harmful inputs (\cref{app:reward_computation}).

\noindent\textbf{Reasoning and General Data.}
To preserve reasoning and general capabilities, we incorporate two additional types of training data. Visual mathematics problems~\cite{ViRL39K} receive verifiable answer rewards~\cite{DeepSeek-R1} to maintain reasoning ability. General instruction data~\cite{LLaVA} receive no explicit reward and are instead anchored to the base model's behavior through the KL constraint in \cref{Joint_Optimization_Objective}. Prior work has shown that incorporating general instruction data into safety fine-tuning mitigates over-refusal while maintaining safety~\cite{VLGuard}.

\subsection{Mode-Coupled Advantage}
\label{Mode_Coupled_Advantage}
Paired rollouts and trace-aware rewards provide safety signals for both modes. However, standard GRPO~\cite{GRPO}, when applied separately to each mode, normalizes rewards within each mode's group, which hides the performance gap between modes and prevents the safer mode from guiding the weaker one. We address this limitation with a shared normalization scale and an explicit cross-mode term.

\noindent\textbf{Shared-Scale Within-Mode Advantage.}
We first compute within-mode advantages using mode-specific baselines and a shared normalization scale. Let $\mu_m$ and $\mu_m^z$ denote the mean answer and trace rewards within mode $m$, respectively, and let $\sigma$ be the standard deviation of the answer rewards of all non-truncated responses in both modes, lower-bounded by $0.1$. The advantages are defined as:
\begin{equation}
A_i=\frac{r_i-\mu_m}{\sigma},\quad
A_i^z=\frac{r_i^z-\mu_m^z}{\sigma}.
\end{equation}
Mode-specific baselines preserve within-mode comparisons, while the shared scale makes advantages comparable across modes. Independent normalization could amplify a mode with predominantly safe responses but low within-group variance, obscuring cross-mode disparity~\cite{Dr-GRPO, REINFORCE++}.

\noindent\textbf{Conditional Cross-Mode Term.}
Extended reasoning can rationalize a harmful request and override the safety behavior a model exhibits when answering directly~\cite{RIM, Self-Jailbreak}. When all ON responses to an input are unsafe, standard GRPO yields zero within-group advantages and no learning signal~\cite{DAPO}, even if the OFF responses to the same input are safe. We therefore use OFF as an in-model reference for ON through one-way cross-mode coupling, gated per input: only when $\mu_{\text{OFF}}>\mu_{\text{ON}}$ do we adjust ON advantages using the OFF group's mean reward as the baseline, while a weaker OFF is up-weighted by the mode weights below:
\begin{equation}
A_i \leftarrow A_i+\gamma\frac{r_i-\mu_{\text{OFF}}}{\sigma},\
A_i^z \leftarrow A_i^z+\gamma\frac{r_i^z-\mu_{\text{OFF}}}{\sigma},
\end{equation}
where $\gamma$ controls the coupling strength ($\gamma=0.5$ by default). Since OFF has no reasoning trace, its mean reward serves as the reference for both ON reward channels. The cross-mode term provides nonzero advantages even when ON rewards are identical, while only responses exceeding the OFF reference receive positive cross-mode contributions. We add a small batch-level correction ($\beta=0.25$)~\cite{REINFORCE++} and clip both advantages to $[-3,3]$ (Supplementary \cref{app:advantage_implementation}).

\noindent\textbf{Segment-Wise Advantage.}
For ON responses, we assign distinct advantages to reasoning and answer tokens:
\begin{equation}
A_{i,t}=\alpha_{m_i}\Big(
\mathds{1}[t\in z_i]\,A_i^z+
\mathds{1}[t\in a_i]\,A_i
\Big),
\end{equation}
while OFF responses contain only answer tokens. An unsafe reasoning trace is thus penalized without suppressing a
safe final answer. The mode weight $\alpha_{\text{ON}}=0.5+0.4\,\operatorname{sigmoid}\big((\mu_{\text{OFF}}-\mu_{\text{ON}})/\tau\big)$, with $\alpha_{\text{OFF}}$ defined symmetrically and $\tau=0.25$, equals $0.7$ for both modes when they perform equally and, akin to group DRO~\cite{DRNN}, allocates more of the update to the mode with the lower mean reward.

\begin{table*}[!t]
\centering
\caption{Cross-Mode Safety Balance (RQ2). $\mathrm{MB}^{m}$ denotes the harmonic mean of $100-\text{ASR}^{m}$ and $100-\text{ORR}^{m}$. WMB is the minimum MB across both modes, while $\Delta\text{ASR}$ measures their absolute ASR gap. Parentheses show percentage-point changes from the base model.}
\setlength{\tabcolsep}{5pt}
\renewcommand{\arraystretch}{1}
\footnotesize
\resizebox{\textwidth}{!}{%
\begin{tabular}{l|cccc|cccc}
\toprule
\multicolumn{1}{c|}{\multirow{2.5}{*}{\textbf{Method}}} &
\multicolumn{4}{c|}{\textbf{USB-Base~\cite{USB}}} &
\multicolumn{4}{c}{\textbf{USB-Hard~\cite{USB}}} \\
\cmidrule(lr){2-5} \cmidrule(l){6-9}
& \textbf{MB$^{\mathrm{OFF}}$ $\uparrow$} & \textbf{MB$^{\mathrm{ON}}$ $\uparrow$} & \textbf{$\Delta$ASR $\downarrow$} & \textbf{WMB $\uparrow$}
& \textbf{MB$^{\mathrm{OFF}}$ $\uparrow$} & \textbf{MB$^{\mathrm{ON}}$ $\uparrow$} & \textbf{$\Delta$ASR $\downarrow$} & \textbf{WMB $\uparrow$} \\
\midrule

\rowcolor{gray!20}
\textbf{Qwen3.5-4B (Base Model)} & 82.9 & 79.0 & 12.0 & 79.0 & 74.4 & 61.5 & 19.7 & 61.5 \\
\; + AdaShield (ECCV'24)~\cite{AdaShield} & \valDown{32.6}{50.3} & \valUp{81.0}{2.0} & \valGapUp{19.4}{7.4} & \valDown{32.6}{46.4} & \valDown{32.6}{41.8} & \valUp{72.7}{11.2} & \valGapUp{32.2}{12.5} & \valDown{32.6}{28.9} \\
\; + ECSO (ECCV'24)~\cite{ECSO} & \valDown{82.6}{0.3} & \valUp{79.4}{0.4} & \valGapDown{11.3}{0.7} & \valUp{79.4}{0.4} & \valUp{75.5}{1.1} & \valUp{66.3}{4.8} & \valGapDown{16.3}{3.4} & \valUp{66.3}{4.8} \\
\; + ETA (ICLR'25)~\cite{ETA} & \valUp{83.3}{0.4} & \valUp{80.2}{1.2} & \valGapUp{12.5}{0.5} & \valUp{80.2}{1.2} & \valUp{75.4}{1.0} & \valUp{63.7}{2.2} & \valGapUp{19.9}{0.2} & \valUp{63.7}{2.2} \\
\; + VLGuard (ICML'24)~\cite{VLGuard} & \valDown{24.9}{58.0} & \valDown{11.9}{67.1} & \valGapDown{4.4}{7.6} & \valDown{11.9}{67.1} & \valDown{24.8}{49.6} & \valDown{11.9}{49.6} & \valGapDown{6.8}{12.9} & \valDown{11.9}{49.6} \\
\; + TiS (EMNLP'25)~\cite{TiS} & \valDown{60.2}{22.7} & \valDown{48.8}{30.2} & \valGapDown{\textbf{1.9}}{\textbf{10.1}} & \valDown{48.8}{30.2} & \valDown{60.5}{13.9} & \valDown{48.9}{12.6} & \valGapDown{\textbf{1.3}}{\textbf{18.4}} & \valDown{48.9}{12.6} \\
\; + MIRage (ICLR'26)~\cite{MIRage} & \valUp{\underline{84.1}}{1.2} & \valDown{76.2}{2.8} & \valGapDown{6.2}{5.8} & \valDown{76.2}{2.8} & \valUp{\textbf{82.1}}{\textbf{7.7}} & \valUp{\underline{76.1}}{14.6} & \valGapDown{10.1}{9.6} & \valUp{\underline{76.1}}{14.6} \\
\; + SafeGRPO (CVPR'26)~\cite{SafeGRPO} & \valDown{81.8}{1.1} & \valUp{\underline{81.1}}{2.1} & \valGapDown{9.5}{2.5} & \valUp{\underline{81.1}}{2.1} & \valUp{76.1}{1.7} & \valUp{68.3}{6.8} & \valGapDown{16.7}{3.0} & \valUp{68.3}{6.8} \\
\rowcolor{skyblue!30}
\; + \textbf{DuoAlign (Ours)} & \valUp{\textbf{86.0}}{\textbf{3.1}} & \valUp{\textbf{86.0}}{\textbf{7.0}} & \valGapDown{\underline{3.0}}{9.0} & \valUp{\textbf{86.0}}{\textbf{7.0}} & \valUp{\underline{81.6}}{7.2} & \valUp{\textbf{81.3}}{\textbf{19.8}} & \valGapDown{\underline{2.8}}{16.9} & \valUp{\textbf{81.3}}{\textbf{19.8}} \\
\midrule

\rowcolor{gray!20}
\textbf{Qwen3.5-9B (Base Model)} & 82.7 & 80.9 & 12.0 & 80.9 & 75.0 & 62.3 & 21.8 & 62.3 \\
\; + AdaShield (ECCV'24)~\cite{AdaShield} & \valDown{48.3}{34.4} & \valUp{\underline{82.4}}{1.5} & \valGapUp{17.0}{5.0} & \valDown{48.3}{32.6} & \valDown{48.0}{27.0} & \valUp{73.9}{11.6} & \valGapUp{27.8}{6.0} & \valDown{48.0}{14.3} \\
\; + ECSO (ECCV'24)~\cite{ECSO} & \valUp{83.2}{0.5} & \valUp{82.2}{1.3} & \valGapDown{11.0}{1.0} & \valUp{\underline{82.2}}{1.3} & \valDown{74.6}{0.4} & \valUp{69.7}{7.4} & \valGapDown{12.6}{9.2} & \valUp{69.7}{7.4} \\
\; + ETA (ICLR'25)~\cite{ETA} & \valUp{83.1}{0.4} & \valUp{81.0}{0.1} & \valGapUp{12.7}{0.7} & \valUp{81.0}{0.1} & \valUp{75.9}{0.9} & \valUp{63.7}{1.4} & \valEq{21.8}{0.0} & \valUp{63.7}{1.4} \\
\; + VLGuard (ICML'24)~\cite{VLGuard} & \valDown{23.9}{58.8} & \valDown{25.6}{55.3} & \valGapDown{3.4}{8.6} & \valDown{23.9}{57.0} & \valDown{23.9}{51.1} & \valDown{25.6}{36.7} & \valGapDown{\textbf{1.8}}{\textbf{20.0}} & \valDown{23.9}{38.4} \\
\; + TiS (EMNLP'25)~\cite{TiS} & \valDown{57.2}{25.5} & \valDown{53.8}{27.1} & \valGapDown{\textbf{0.5}}{\textbf{11.5}} & \valDown{53.8}{27.1} & \valDown{57.1}{17.9} & \valDown{54.1}{8.2} & \valGapDown{\underline{3.4}}{18.4} & \valDown{54.1}{8.2} \\
\; + MIRage (ICLR'26)~\cite{MIRage} & \valUp{\textbf{84.6}}{\textbf{1.9}} & \valDown{79.6}{1.3} & \valGapDown{\underline{2.2}}{9.8} & \valDown{79.6}{1.3} & \valUp{\textbf{82.8}}{\textbf{7.8}} & \valUp{\underline{79.5}}{17.2} & \valGapDown{5.7}{16.1} & \valUp{\underline{79.5}}{17.2} \\
\; + SafeGRPO (CVPR'26)~\cite{SafeGRPO} & \valDown{82.0}{0.7} & \valDown{80.4}{0.5} & \valGapDown{11.2}{0.8} & \valDown{80.4}{0.5} & \valUp{75.7}{0.7} & \valUp{66.0}{3.7} & \valGapDown{18.7}{3.1} & \valUp{66.0}{3.7} \\
\rowcolor{skyblue!30}
\; + \textbf{DuoAlign (Ours)} & \valUp{\underline{84.2}}{1.5} & \valUp{\textbf{87.2}}{\textbf{6.3}} & \valGapDown{4.0}{8.0} & \valUp{\textbf{84.2}}{\textbf{3.3}} & \valUp{\underline{81.0}}{6.0} & \valUp{\textbf{83.7}}{\textbf{21.4}} & \valGapDown{3.8}{18.0} & \valUp{\textbf{81.0}}{\textbf{18.7}} \\
\midrule

\rowcolor{gray!20}
\textbf{Gemma-4-E4B (Base Model)} & 79.5 & 80.2 & 5.6 & 79.5 & 72.6 & 65.0 & 14.3 & 65.0 \\
\; + AdaShield (ECCV'24)~\cite{AdaShield} & \valDown{74.4}{5.1} & \valUp{81.1}{0.9} & \valGapDown{5.3}{0.3} & \valDown{74.4}{5.1} & \valDown{72.2}{0.4} & \valUp{70.4}{5.4} & \valGapUp{16.9}{2.6} & \valUp{70.4}{5.4} \\
\; + ECSO (ECCV'24)~\cite{ECSO} & \valUp{79.6}{0.1} & \valUp{81.3}{1.1} & \valGapDown{4.5}{1.1} & \valUp{79.6}{0.1} & \valUp{74.2}{1.6} & \valUp{66.5}{1.5} & \valGapUp{15.4}{1.1} & \valUp{66.5}{1.5} \\
\; + ETA (ICLR'25)~\cite{ETA} & \valUp{79.8}{0.3} & \valUp{80.7}{0.5} & \valGapUp{5.9}{0.3} & \valUp{79.8}{0.3} & \valUp{74.4}{1.8} & \valUp{65.7}{0.7} & \valGapUp{16.9}{2.6} & \valUp{65.7}{0.7} \\
\; + VLGuard (ICML'24)~\cite{VLGuard} & \valDown{37.1}{42.4} & \valDown{41.5}{38.7} & \valGapDown{5.1}{0.5} & \valDown{37.1}{42.4} & \valDown{37.0}{35.6} & \valDown{41.5}{23.5} & \valGapDown{7.0}{7.3} & \valDown{37.0}{28.0} \\
\; + TiS (EMNLP'25)~\cite{TiS} & \valDown{54.6}{24.9} & \valDown{52.8}{27.4} & \valGapDown{3.1}{2.5} & \valDown{52.8}{26.7} & \valDown{54.8}{17.8} & \valDown{52.8}{12.2} & \valGapDown{\textbf{1.6}}{\textbf{12.7}} & \valDown{52.8}{12.2} \\
\; + MIRage (ICLR'26)~\cite{MIRage} & \valUp{\underline{80.6}}{1.1} & \valUp{\underline{83.6}}{3.4} & \valGapUp{8.5}{2.9} & \valUp{\underline{80.6}}{1.1} & \valUp{\textbf{79.7}}{\textbf{7.1}} & \valUp{\textbf{83.3}}{\textbf{18.3}} & \valGapDown{9.6}{4.7} & \valUp{\textbf{79.7}}{\textbf{14.7}} \\
\; + SafeGRPO (CVPR'26)~\cite{SafeGRPO} & \valDown{78.6}{0.9} & \valUp{80.8}{0.6} & \valGapDown{\underline{2.8}}{2.8} & \valDown{78.6}{0.9} & \valUp{74.2}{1.6} & \valUp{69.4}{4.4} & \valGapDown{11.5}{2.8} & \valUp{69.4}{4.4} \\
\rowcolor{skyblue!30}
\; + \textbf{DuoAlign (Ours)} & \valUp{\textbf{81.6}}{\textbf{2.1}} & \valUp{\textbf{84.4}}{\textbf{4.2}} & \valGapDown{\textbf{1.7}}{\textbf{3.9}} & \valUp{\textbf{81.6}}{\textbf{2.1}} & \valUp{\underline{79.3}}{6.7} & \valUp{\underline{79.4}}{14.4} & \valGapDown{\underline{3.1}}{11.2} & \valUp{\underline{79.3}}{14.3} \\
\bottomrule
\end{tabular}%
}
\label{Exp_RQ2_Table}
\end{table*}

\subsection{Joint Optimization Objective}
\label{Joint_Optimization_Objective}

We jointly optimize both modes using a unified clipped surrogate objective~\cite{PPO}, with the base model as the reference policy $\pi_{\text{ref}}$. Let $\bar{\rho}_{i,t}=\operatorname{clip}(\rho_{i,t},1-\epsilon,1+\epsilon)$:
\begin{align}
\mathcal{J}(\theta)
&=\mathbb{E}\Big[
\sum_x\sum_{y_i\in\mathcal{G}(x)}\Big(\sum_t
w_{i,t}\min(\rho_{i,t}A_{i,t},\bar{\rho}_{i,t}A_{i,t})\notag\\
&\qquad-\kappa_x D_{\mathrm{KL}}^{(i)}\Big)\Big],
\label{eq:joint}
\end{align}
where $\rho_{i,t}$ denotes the token-level probability ratio between the new and old policies and $\epsilon=0.2$. The token weight $w_{i,t}$ gives the reasoning and answer segments of an ON response equal total weight at fixed reference lengths (step-0 medians), and is zero for truncated responses. We employ a low-variance token-level KL estimator $D_{\mathrm{KL}}^{(i)}$ against $\pi_{\text{ref}}$ over all valid sampled responses, with $\kappa_x=0.01$ for safety and visual-math inputs and a larger $\kappa_x=0.05$ for general instruction data to preserve general capabilities (\cref{app:optimization_details}).

\noindent\textbf{Two-Stage Training.}
As training progresses, more paired groups become uniformly safe, so their rewards saturate and advantages vanish~\cite{DAPO}. We therefore train in two stages with hard example mining~\cite{OHEM}. Specifically, Stage 1 retains a harmful input only if at least one of the base model's $2K$ responses to it is unsafe. Stage 2 then resumes from the final Stage-1 checkpoint on the other half of the training data, keeping its benign, visual-math, and general-instruction inputs and re-mining its harmful inputs with the Stage-1 policy. Unlike per-step dynamic sampling in DAPO~\cite{DAPO}, mining only once per stage keeps the overhead small.
\section{Experiments}
\label{Experiments}

\subsection{Experimental Setup}

\textbf{Base Models.}
We conduct experiments on three representative hybrid-thinking VLMs: Qwen3.5-4B/9B~\cite{Qwen3.5} and Gemma-4-E4B~\cite{Gemma4}. DuoAlign trains rank-16 LoRA~\cite{LoRA} adapters on the language backbone with verl~\cite{verl}, keeping the visual encoder frozen, and we report the checkpoint with the best safety performance on a held-out development set.

\begin{figure*}[t]
    \centering
    \includegraphics[width=\linewidth]{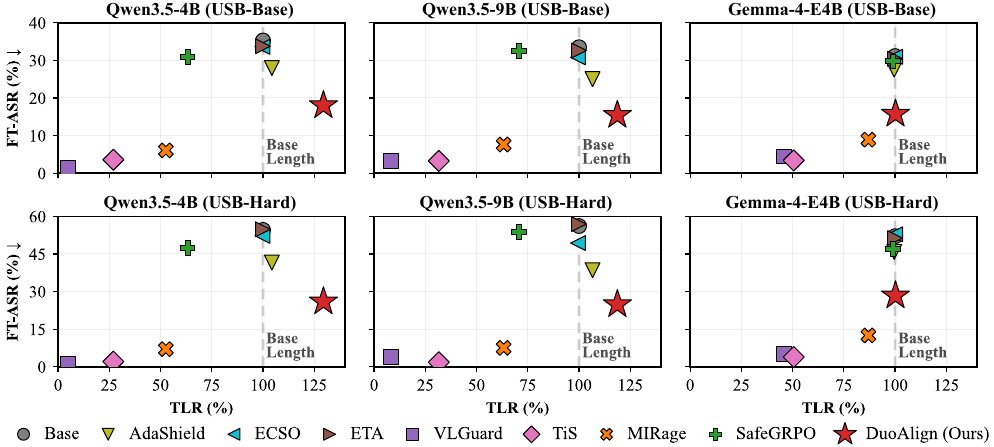}
    \caption{Reasoning Preservation (RQ3). FT-ASR is measured on USB-Base (top) and USB-Hard (bottom). TLR is measured on MOSSBench and shared between the two rows. Lower right is better: low FT-ASR with TLR at or above the base reasoning length (dashed line).}
    \label{Exp_RQ3_Figure}
\end{figure*}

\noindent\textbf{Training Data.}
Our training set comprises 4,000 examples: 3,000 safety samples (1,500 harmful and 1,500 benign) from SPA-VL~\cite{SPA-VL} and BeaverTails-V~\cite{BeaverTails}, 700 visual mathematics problems from ViRL39K~\cite{ViRL39K}, and 300 general instructions from LLaVA-Instruct~\cite{LLaVA}. Training proceeds in two successive stages of 2,000 examples each, with harmful samples re-mined in Stage 2. We remove training-development overlaps and deduplicate all safety training data, including baseline data, against all USB~\cite{USB} and MOSSBench~\cite{MOSSBench} data using both image and text similarity. General training data are further deduplicated against the five capability benchmarks~\cite{MathVista, MMMU, MM-Vet, MMBench, POPE}.

\noindent\textbf{Baselines.}
We compare DuoAlign against seven representative baselines across three categories: training-free defenses, including AdaShield~\cite{AdaShield}, ECSO~\cite{ECSO}, and ETA~\cite{ETA}; supervised fine-tuning methods, including VLGuard~\cite{VLGuard}, TiS~\cite{TiS}, and MIRage~\cite{MIRage}; and reinforcement learning, represented by SafeGRPO~\cite{SafeGRPO}. We adapt all baselines to the same three backbones and consistently evaluate both modes under a shared protocol (Supplementary \cref{app:evaluation_details}).

\noindent\textbf{Evaluation Benchmarks.}
We adopt USB~\cite{USB} to evaluate safety, which integrates 13 established safety datasets~\cite{Hades, HarmBench, VLSafe, JailBreakV, MLLMGuard, MM-SafetyBench, MSSBench, MultiTrust, RTVLM, SafeBench, VLGuard, VLSBench, MOSSBench}. Across 16 subcategories and four image-text safety types, we sample 24 examples per category-type combination from USB-Base (1,536 samples) and 6 from USB-Hard (384 samples). We evaluate over-refusal on MOSSBench~\cite{MOSSBench} and general capabilities on MathVista~\cite{MathVista}, MMMU~\cite{MMMU}, MM-Vet~\cite{MM-Vet}, MMBench~\cite{MMBench}, and POPE~\cite{POPE}. We use VLMEvalKit~\cite{VLMEvalKit} to standardize general multimodal capability evaluation.

\noindent\textbf{Evaluation Protocol.}
We adopt all metrics defined in \cref{Preliminaries}. Each method generates one response per mode using the model-recommended sampling configuration. Evaluation rates are computed for each method and metric over successfully judged samples with valid output formats, excluding the few failed or invalid cases. GPT-6 Luna~\cite{Luna} serves as the primary safety and refusal judge, independently of the DeepSeek-V4.1-Flash-1017~\cite{DeepSeek1017} training reward judge. To mitigate single-judge bias, we additionally conduct cross-evaluations with Claude Haiku 5.5~\cite{Haiku}, obtaining consistent overall results (Supplementary \cref{Cross_Judge}).


\subsection{Safety--Helpfulness Balance (RQ1)}
\label{Exp_RQ1}
DuoAlign achieves the best Safety--Helpfulness balance across all three models (\cref{Exp_RQ1_Table}). On USB-Hard~\cite{USB}, it improves HM(DSR, DAR) from 55.1\%/55.3\%/58.0\% to 74.0\%/74.5\%/72.5\%, yielding gains of 18.9/19.2/14.5 pp. On USB-Base~\cite{USB}, DuoAlign also achieves the highest HM across all three models, reaching 79.9\%/79.1\%/77.0\%. Notably, safety improvements often come at the cost of increased over-refusal. SFT-based methods achieve high DSR but suffer from substantial over-refusal~\cite{VLGuard, TiS, MIRage}, while other baselines provide relatively limited safety gains. In contrast, DuoAlign reduces DAR by only 4.9--7.6 pp relative to the base models and maintains a dual-mode answer rate of 99.0--100\% on the MSSBench~\cite{MSSBench} over-refusal benchmark (\cref{Apendix_MSSBench}). DuoAlign thus sharpens the response boundary on risky inputs instead of refusing broadly.

\subsection{Cross-Mode Safety Balance (RQ2)}
\label{Exp_RQ2}
As shown in \cref{Exp_RQ2_Table}, DuoAlign substantially narrows cross-mode safety gaps while improving the weaker mode. On USB-Hard~\cite{USB}, DuoAlign reduces $\Delta\text{ASR}$ from 19.7/21.8/14.3 to 2.8/3.8/3.1 pp, while increasing WMB from 61.5/62.3/65.0\% to 81.3/81.0/79.3\% across the three models, respectively. DuoAlign achieves the highest WMB on Qwen3.5-4B/9B and is within 0.4 pp of the best method (MIRage) on Gemma-4-E4B. On USB-Base~\cite{USB}, it achieves the highest WMB across all three models. In comparison, TiS~\cite{TiS} and VLGuard~\cite{VLGuard} sometimes attain smaller $\Delta\text{ASR}$ but suffer substantial over-refusal, reducing WMB to 11.9--54.1\%. Training-free defenses instead widen the gap in six of nine USB-Hard settings, with AdaShield widening it by 12.5 pp on Qwen3.5-4B (19.7 to 32.2 pp). These improvements primarily stem from strengthening the weaker ON mode. On USB-Hard, DuoAlign improves $\text{MB}^{\mathrm{ON}}$ by 19.8/21.4/14.4 pp over the base models, while also improving $\text{MB}^{\mathrm{OFF}}$ by 7.2/6.0/6.7 pp. Thus, the reduced safety gap reflects stronger ON protection without compromising OFF performance. Beyond aggregate metrics, we further examine instance-level consistency (Supplementary \cref{app:instance_safety}). DuoAlign reduces disagreement from 29.2/30.8/24.7\% to 18.3/16.5/16.1\% while increasing dual-mode safe rates by 27.1/31.5/24.3 pp. The remaining disagreement is also far more symmetric: ON-only unsafe cases drop from 24.5/26.4/19.5\% to 10.9/10.2/9.6\%, close to the OFF-only rates of 7.4/6.3/6.5\%. Improved cross-mode safety balance thus accompanies stronger joint safety rather than shared unsafe behavior.

\begin{table*}[!t]
\centering
\caption{Capability Preservation (RQ4). Average score over five benchmarks. Parentheses show changes from the base model.}

\setlength{\tabcolsep}{4pt}
\renewcommand{\arraystretch}{1.05}
\footnotesize
\resizebox{\textwidth}{!}{%
\begin{tabular}{ll|c|ccccccc|c}
\toprule
\multicolumn{1}{c}{\textbf{Model}} & \multicolumn{1}{c|}{\textbf{Mode}} &
\cellcolor{gray!15}\textbf{Base} & \textbf{AdaShield~\cite{AdaShield}} & \textbf{ECSO~\cite{ECSO}} & \textbf{ETA~\cite{ETA}} &
\textbf{VLGuard~\cite{VLGuard}} & \textbf{TiS~\cite{TiS}} & \textbf{MIRage~\cite{MIRage}} & \textbf{SafeGRPO~\cite{SafeGRPO}} &
\cellcolor{skyblue!30}\textbf{DuoAlign} \\
\midrule

\multirow{2}{*}{\textbf{Qwen3.5-4B}}
& ON & \cellcolor{gray!15}\underline{80.3} & \valDown{79.9}{0.4} & \valDown{80.0}{0.3} & \valDown{79.8}{0.5} & \valDown{69.0}{11.3} & \valDown{73.2}{7.1} & \valDown{79.4}{0.9} & \valUp{\textbf{80.7}}{\textbf{0.4}} & \cellcolor{skyblue!30}\valEq{\underline{80.3}}{0.0} \\
& OFF & \cellcolor{gray!15}\underline{78.7} & \valDown{78.0}{0.7} & \valDown{78.4}{0.3} & \valDown{78.2}{0.5} & \valDown{70.6}{8.1} & \valDown{77.7}{1.0} & \valDown{73.0}{5.7} & \valDown{78.6}{0.1} & \cellcolor{skyblue!30}\valUp{\textbf{78.8}}{\textbf{0.1}} \\
\midrule

\multirow{2}{*}{\textbf{Qwen3.5-9B}}
& ON & \cellcolor{gray!15}\underline{82.5} & \valDown{82.0}{0.5} & \valDown{82.2}{0.3} & \valDown{82.1}{0.4} & \valDown{74.3}{8.2} & \valDown{76.0}{6.5} & \valDown{82.2}{0.3} & \valDown{82.2}{0.3} & \cellcolor{skyblue!30}\valUp{\textbf{82.7}}{\textbf{0.2}} \\
& OFF & \cellcolor{gray!15}\underline{81.0} & \valDown{80.7}{0.3} & \valDown{80.8}{0.2} & \valEq{\underline{81.0}}{0.0} & \valDown{73.8}{7.2} & \valDown{79.6}{1.4} & \valDown{74.8}{6.2} & \valDown{80.9}{0.1} & \cellcolor{skyblue!30}\valUp{\textbf{81.2}}{\textbf{0.2}} \\
\midrule

\multirow{2}{*}{\textbf{Gemma-4-E4B}}
& ON & \cellcolor{gray!15}68.8 & \valUp{\underline{69.7}}{0.9} & \valUp{69.4}{0.6} & \valUp{69.5}{0.7} & \valDown{63.8}{5.0} & \valDown{62.7}{6.1} & \valDown{68.1}{0.7} & \valUp{\textbf{69.8}}{\textbf{1.0}} & \cellcolor{skyblue!30}\valUp{69.0}{0.2} \\
& OFF & \cellcolor{gray!15}\textbf{68.4} & \valEq{\textbf{68.4}}{0.0} & \valDown{\underline{68.2}}{0.2} & \valDown{68.1}{0.3} & \valDown{61.2}{7.2} & \valDown{66.3}{2.1} & \valDown{65.4}{3.0} & \valDown{68.1}{0.3} & \cellcolor{skyblue!30}\valDown{67.2}{1.2} \\
\bottomrule
\end{tabular}%
}
\label{Exp_RQ4_Table}
\end{table*}

\subsection{Reasoning Preservation (RQ3)}
\label{Exp_RQ3}
DuoAlign makes reasoning safer without suppressing it (\cref{Exp_RQ3_Figure}). On USB-Base~\cite{USB}, DuoAlign reduces FT-ASR from 35.3\%/33.5\%/31.1\% to 18.0\%/15.4\%/15.8\%, while maintaining TLRs of 130\%/119\%/100\%. Among all compared methods, DuoAlign is the only approach that achieves both low reasoning risk and preserved reasoning length. Although VLGuard, TiS, and MIRage achieve lower FT-ASR, their TLRs drop substantially, indicating significant reasoning suppression. Training-free defenses and SafeGRPO retain more reasoning length but exhibit much higher FT-ASR (25.0--33.8\%). The advantage grows under harder attacks: on USB-Hard, DuoAlign reduces FT-ASR from 54.6\%/56.1\%/52.1\% to 26.0\%/24.9\%/28.4\%, larger reductions (28.6/31.2/23.7 pp) than on USB-Base (17.3/18.1/15.3 pp), and its margin over the best length-preserving baseline widens from 9.6--11.7 to 13.6--17.4 pp. Reasoning quality is preserved as well, as the ON-mode capability scores remain unchanged or slightly improve (\cref{Exp_RQ4}).

\subsection{Capability Preservation (RQ4)}
\label{Exp_RQ4}
DuoAlign preserves the general capabilities of both modes across five benchmarks (\cref{Exp_RQ4_Table}). On Qwen3.5-4B and Qwen3.5-9B~\cite{Qwen3.5}, the average benchmark score of both modes remains unchanged or slightly improves. On Gemma-4-E4B~\cite{Gemma4}, the average score improves by 0.2 points in ON and decreases by 1.2 points in OFF. In contrast, several SFT baselines incur substantially larger performance degradation. VLGuard~\cite{VLGuard} reduces the average score by 5.0--11.3 points across both modes, TiS~\cite{TiS} by 6.1--7.1 points in ON, and MIRage~\cite{MIRage} by 3.0--6.2 points in OFF. SafeGRPO~\cite{SafeGRPO} and training-free defenses generally keep changes in the average score within 1.0 point, but yield considerably smaller safety gains (RQ1--RQ3). Per-benchmark results are given in Supplementary \cref{RQ4_Complete}.

\begin{table}[!t]
\centering
\caption{Ablations on Qwen3.5-4B. Safety metrics use USB-Hard and MOSSBench, while Score averages capability across both modes. For fair comparison, all variants, including Full, use their final checkpoints. Parentheses indicate changes relative to Full.}
\label{Exp_AB_Table}
\setlength{\tabcolsep}{3pt}
\renewcommand{\arraystretch}{1.05}
\footnotesize
\resizebox{\columnwidth}{!}{%
\begin{tabular}{l|cccc}
\toprule
\multicolumn{1}{c|}{\multirow{2}{*}{\textbf{Variant}}} &
\textbf{HM $\uparrow$} & \textbf{WMB $\uparrow$} &
\textbf{FT-ASR $\downarrow$} & \textbf{Score $\uparrow$} \\
& \textbf{(RQ1)} & \textbf{(RQ2)} & \textbf{(RQ3)} & \textbf{(RQ4)} \\
\midrule
\rowcolor{gray!10}
& \multicolumn{4}{c}{\textbf{Mode Optimization}} \\
Independent-Mode & \valDown{70.6}{4.6} & \valDown{78.4}{4.6} & \valGapUp{32.6}{5.6} & \valDown{79.7}{0.1} \\
Pooled-GRPO & \valDown{72.3}{2.9} & \valDown{81.1}{1.9} & \valGapUp{32.1}{5.1} & \valDown{79.5}{0.3} \\
ON-only & \valDown{70.2}{5.0} & \valDown{78.3}{4.7} & \valGapUp{30.2}{3.2} & \valDown{79.0}{0.8} \\
\midrule
\rowcolor{gray!10}
& \multicolumn{4}{c}{\textbf{Reward and Cross-Mode}} \\
\begin{tabular}[c]{@{}l@{}}
Leakage-Only +\\[-2pt]No Cross-Mode
\end{tabular}
& \valDown{73.2}{2.0} & \valDown{80.9}{2.1} & \valGapUp{30.2}{3.2} & \valEq{\underline{79.8}}{0.0} \\
Leakage-Only & \valDown{\underline{75.0}}{0.2} & \valDown{\underline{82.1}}{0.9} & \valGapDown{\textbf{24.7}}{2.3} & \valDown{79.3}{0.5} \\
No Cross-Mode & \valDown{74.7}{0.5} & \valDown{81.8}{1.2} & \valGapUp{28.1}{1.1} & \valUp{\textbf{79.9}}{0.1} \\
\midrule
\rowcolor{skyblue!30}
\textbf{DuoAlign (Full)} & \textbf{75.2} & \textbf{83.0} & \underline{27.0} & \underline{79.8} \\
\bottomrule
\end{tabular}%
}
\end{table}
\vspace{-0.3cm}

\subsection{Ablation Studies}
\label{Exp_AB}

\noindent\textbf{Mode-Optimization Configurations.}
We investigate three representative training variants: \textit{Independent-Mode} normalizes each mode with its own scale, fixes both mode weights to $0.7$, and removes coupling, \textit{Pooled-GRPO} jointly normalizes ON/OFF rewards without mode-specific baselines or conditional coupling, and \textit{ON-only} trains exclusively on ON responses, following the single-mode practice of prior work. Compared with DuoAlign, they reduce HM by 4.6/2.9/5.0 pp and WMB by 4.6/1.9/4.7 pp, while increasing FT-ASR by 5.6/5.1/3.2 pp (\cref{Exp_AB_Table}). ON-only yields the lowest HM and WMB of all variants, confirming that aligning ON alone does not secure both modes. Pooling improves over independent optimization but remains inferior to DuoAlign, showing the clear benefit of mode-aware advantage estimation beyond naive pooling.

\noindent\textbf{Reward and Cross-Mode Configurations.}
\textit{Leakage-Only} replaces the full-output safety judgment for harmful ON responses with a reasoning-only leakage penalty of 0.4, \textit{No Cross-Mode} sets $\gamma=0$, and \textit{Leakage-Only + No Cross-Mode} combines both. Each change lowers HM and WMB, and removing both costs the most (2.0 pp HM, 2.1 pp WMB, and +3.2 pp FT-ASR). Leakage-Only attains a lower FT-ASR (24.7\% vs.\ 27.0\%) but trails DuoAlign on HM and WMB, the two metrics that require safety and responsiveness jointly. General capabilities remain stable across all variants. Further, comparing Independent-Mode with No Cross-Mode shows that the shared scale and adaptive mode weights together account for most of the HM and WMB gains (4.1 and 3.4 pp), while the cross-mode term adds smaller but consistent further gains (0.5 pp HM and 1.2 pp WMB) and lowers FT-ASR by 1.1 pp.
\section{Conclusion}
Hybrid thinking gives one model two ways to answer, and we show that the two are not equally safe: the ON mode is consistently more vulnerable, many harmful inputs receive conflicting outcomes across modes, and safe-looking answers can hide harmful reasoning. We formalize four essential safety criteria for this setting and introduce DuoAlign, which aligns both modes jointly through Paired-Mode Rollout, Trace-Aware Reward, and Mode-Coupled Advantage. Across three models, DuoAlign achieves the best safety--helpfulness balance, closes most of the cross-mode safety gap, makes reasoning safer without suppressing it, and effectively preserves general multimodal capabilities.

\section*{LLM Usage Statement}
DeepSeek-V4.1-Flash-1017~\cite{DeepSeek1017} served as the training reward judge and was used for safety data selection and hard-example mining (\cref{Method}; Supplementary \cref{Implementation_Details}). GPT-6 Luna~\cite{Luna} served as the primary evaluation judge, and Claude Haiku 5.5~\cite{Haiku} as an independent cross-judge (\cref{Experiments}; Supplementary~\cref{Cross_Judge}). ChatGPT assisted with code review and evaluation-script development. Claude was used to review the manuscript and suggest wording revisions. The authors designed the research, verified all code, results, and text, and take full responsibility for the manuscript.
{
    \small
    \bibliographystyle{ieeenat_fullname}
    \bibliography{main}
}

\end{document}